\documentclass[conference]{IEEEtran}
\IEEEoverridecommandlockouts
\usepackage{cite}
\usepackage{amsmath,amssymb,amsfonts}
\usepackage{algorithmic}
\usepackage{graphicx}
\usepackage{textcomp}
\usepackage{xcolor}
\def\BibTeX{{\rm B\kern-.05em{\sc i\kern-.025em b}\kern-.08em    
    T\kern-.1667em\lower.7ex\hbox{E}\kern-.125emX}}
\begin{document}

\title{RaycastGrasp: Eye-Gaze Interaction with Wearable Devices for Robotic Manipulation \\

}

\author{\IEEEauthorblockN{First A. Zitiantao Lin and Yongpeng Sang.  \textit {Member, IEEE}}
}

\author{
\IEEEauthorblockN{Zitiantao Lin\textsuperscript{*}}
\IEEEauthorblockA{
College of Engineering\\
Northeastern University\\
Boston, USA\\
lin.ziti@northeastern.edu}
\and
\IEEEauthorblockN{Yongpeng Sang\textsuperscript{*}}
\IEEEauthorblockA{
Khoury College of Computer Sciences\\
Northeastern University\\
Boston, USA\\
sang.yo@northeastern.edu}
\and
\IEEEauthorblockN{Yang Ye}
\IEEEauthorblockA{
Department of Civil \& Environmental Engineering\\
Northeastern University\\
Boston, USA\\
y.ye@northeastern.edu}
\thanks{\textsuperscript{*}These authors contributed equally to this work.}
}

\maketitle

\begin{abstract}
Robotic manipulators are increasingly used to assist individuals with mobility impairments in object retrieval. However, the predominant joystick-based control interfaces can be challenging due to high precision requirements and unintuitive reference frames. Recent advances in human-robot interaction have explored alternative modalities, yet many solutions still rely on external screens or restrictive control schemes, limiting their intuitiveness and accessibility. To address these challenges, we present an egocentric, gaze-guided robotic manipulation interface that leverages a wearable Mixed Reality (MR) headset. Our system enables users to interact seamlessly with real-world objects using natural gaze fixation from a natural first-person perspective, while providing augmented visual cues to confirm intent and leveraging a pretrained vision model and robotic arm for intent recognition and object manipulation. Experimental results demonstrate that our approach significantly improves manipulation accuracy, reduces system latency, and achieves single‑pass intention and objection recognition accuracy $>$ 88\% across multiple real-world scenarios. These results demonstrate the system's effectiveness in enhancing intuitiveness and accessibility, underscoring its practical significance for assistive robotics applications.
\end{abstract}


\section{Introduction}
Although joystick-based control remains the most common approach for wheelchair robots and industrial manipulators\cite{lambrecht2021joystick}  \cite{pawlowski2022comparison}, this method has notable drawbacks, including high precision requirements, unintuitive reference frames, and substantial cognitive and physical demands on users. In practice, users must repeatedly adjust joystick direction and force, which can lead to fatigue and reduced operational efficiency—especially for individuals with limited hand dexterity or motor impairments.

To address these challenges, recent years have seen the emergence of alternative control methods such as reference frame reallocation\cite{okamura2021assistive}, gesture control, and vision-based interfaces. However, these solutions often require extensive training, lack intuitive feedback, or suffer from reliability issues\cite{lambrecht2021joystick}. Gaze-based robotic interaction has thus emerged as a promising alternative, enabling intuitive, hands-free operation through natural gaze behavior.

Currently, various eye-gaze-based control systems\cite{vargas2021headfree} have been developed within the Human-Robot Interaction (HRI) framework\cite{parg2021handgesture}. One of the earliest applications utilizing eye tracking for creative tasks was EagleEyes\cite{scalera2021eyedrawing}, a control system enabling individuals with disabilities to draw on a screen. Another example is the GazeGrasp system\cite{tokmurziyev2025gazegrasp}, which enables robotic grasping by interpreting users’ gaze on a computer screen. However, most of these systems rely on a third-party screen as the interaction medium, using the robot's perspective as the primary viewpoint for manipulation. Alternatively, some emerging eye-gaze models employ head-mounted eye trackers, but these approaches often constrain the user's head posture or rely on supplementary optical tracking devices, such as OptiTrack, to accurately capture head movements.

To overcome these challenges, our approach emphasizes egocentric, attention-based interaction, enabling users to intuitively select and manipulate real-world objects through natural gaze fixation in a Mixed Reality (MR) environment. By leveraging gaze as an explicit indicator of user intent, the system supports hands-free, first-person HRI.

To realize this, we design a system that seamlessly integrates a Franka Emika collaborative robotic arm, a wearable MR headset, and a self-developed eye-tracking algorithm. The user's gaze, captured from their own perspective, is used to identify and select objects in the environment. Object recognition is performed via a pretrained YOLOv8 deep learning model, while the robotic arm executes the corresponding manipulation. This design bridges immersive, user-centered MR experiences with practical robotic assistance, showcasing the power of intuitive, gaze-driven control. Experimental results show that this approach significantly improves manipulation accuracy, reduces system latency, and enhances user satisfaction compared to traditional screen-based eye-tracking methods. The system achieves a single‑pass intention and objection recognition accuracy confidence of $>$ 88\% across multiple real-world scenarios, demonstrating its robustness and versatility.

\section{Related Work}

\subsection{Gaze Fixation as a Cognitive and Social Signal}
Gaze fixation, defined as the sustained visual attention directed at a particular object or spatial location, serves as a crucial indicator of both cognitive processing and social intention [ref]. In instructional contexts, teacher gaze patterns have been shown to reflect specific pedagogical goals: fixating on students is often used to guide attention, on instructional content to emphasize key information, and on blank spaces to indicate cognitive effort such as planning or hesitation [ref]. These patterns suggest that both the duration and spatial target of a fixation can reveal an individual's cognitive load and communicative intent\cite{baranwal2021teachergaze}.

In the domain of HRI, fixation duration plays a similarly critical role in how users interpret the robot's intent. Short glances (less than 1 second) are typically perceived as unintentional or non-interactive, while prolonged fixations (exceeding 2 seconds) are more likely interpreted as signals of interest, inquiry, or social engagement\cite{loth2021roboticgaze}. These findings highlight the importance of modeling fixation duration when designing gaze-based interactive systems.

Beyond temporal features, gaze fixation is also central to semantic inference and intention understanding. Neurophysiological evidence shows that observing another’s gaze toward an object activates brain regions akin to those triggered by observing goal-directed actions like reaching or grasping. This supports the view that gaze serves as a non-verbal proxy for intention communication \cite{goodwin2021socialgaze}.

Fixation has also been shown to strongly correlate with attention allocation and cognitive effort. For example, in natural tasks, individuals tend to fixate on objects before acting on them, and shift gaze only once the action is nearly complete\cite{goodwin2021socialgaze}. Moreover, gaze aversion—momentarily looking away—is often employed to regulate cognitive processing or reduce interpersonal stress, depending on the duration and context of the fixation\cite{goodwin2021socialgaze}.

This behavior and neurological studies provide solid theoretical scaffolding for gaze-based intention prediction that can be applied in assistive robots.

\subsection{Gaze-Based Robotic Interaction in MR and HRI}
Gaze behavior is a crucial interaction modality in both MR and HRI, as it reflects the user’s real-time focus of attention and underlying intention. In MR environments, Baptista et al.~\cite{baptista2025mihrage} proposed an interaction system called MIHRaGe, which enables users with limited limb mobility to control a robotic arm solely through eye movements. When a user fixates on a specific point on the interface, a menu with four options—MOVE, PICK, PLACE, and CANCEL—is triggered, allowing for contactless robotic control. The system achieves a task success rate of 80\%, using a 2-second fixation as the selection trigger. However, the system requires that all target objects be pre-placed within a predefined workspace, which limits its adaptability to real-world environments.

To address this limitation, Wang et al.~\cite{wang2023what} developed a gaze-driven robotic grasping system that combines near-eye tracking with GraspFormer, a transformer-based grasp candidate generator. By estimating 3D gaze vectors and aligning them with generated grasp candidates, the system achieves over 95\% grasping accuracy with a response latency below 100~ms. However, as the interaction still occurs on a 2D image plane, it lacks true MR spatial anchoring, which may hinder depth perception and reduce interaction naturalness.

Furthermore, Choi et al.~\cite{choi2024gaze} introduced an MR-based HRI system for infrastructure inspection that classifies gaze behavior into scanning, focusing, and inspecting categories. It employs a virtual drone for spatial feedback, dynamically adjusting its position based on real-time gaze data. Although the system is not designed for robotic manipulation and does not incorporate object recognition, it demonstrates the potential of gaze-based attention tracking in dynamic environments.

In addition, Li et al.~\cite{zhang2022accurate} designed a binocular eye-tracking system for controlling a 7-DoF anthropomorphic robotic arm. Their approach integrates real-time 3D gaze estimation with anatomical modeling and corneal refraction correction to improve targeting precision. The system also incorporates dynamic inverse kinematics compensation to account for head movement, enabling accurate gaze-controlled manipulation across various tasks, such as cup stacking and peg insertion. This study confirms that 3D gaze serves as a reliable and high-precision input modality for complex robotic operations.

\section{Method/System design}

\subsection{System/Architecture Overview}
Through a review of existing gaze-tracking-based robotic manipulation systems, we observe that most current approaches rely on images captured from the robot's perspective to perform object localization and grasping [ref], which either constrain the motion of users or rely on additional screens for display. 
However, true Mixed Reality (MR) scenarios require the user’s gaze to interact directly with real-world objects, rather than merely analyzing static images on a computer screen. To address this gap, our system leverages an MR headset to enable users to observe and select real objects from their own perspective using eye gaze.
\begin{figure}[htbp]
    \centering
    \includegraphics[width=1\linewidth]{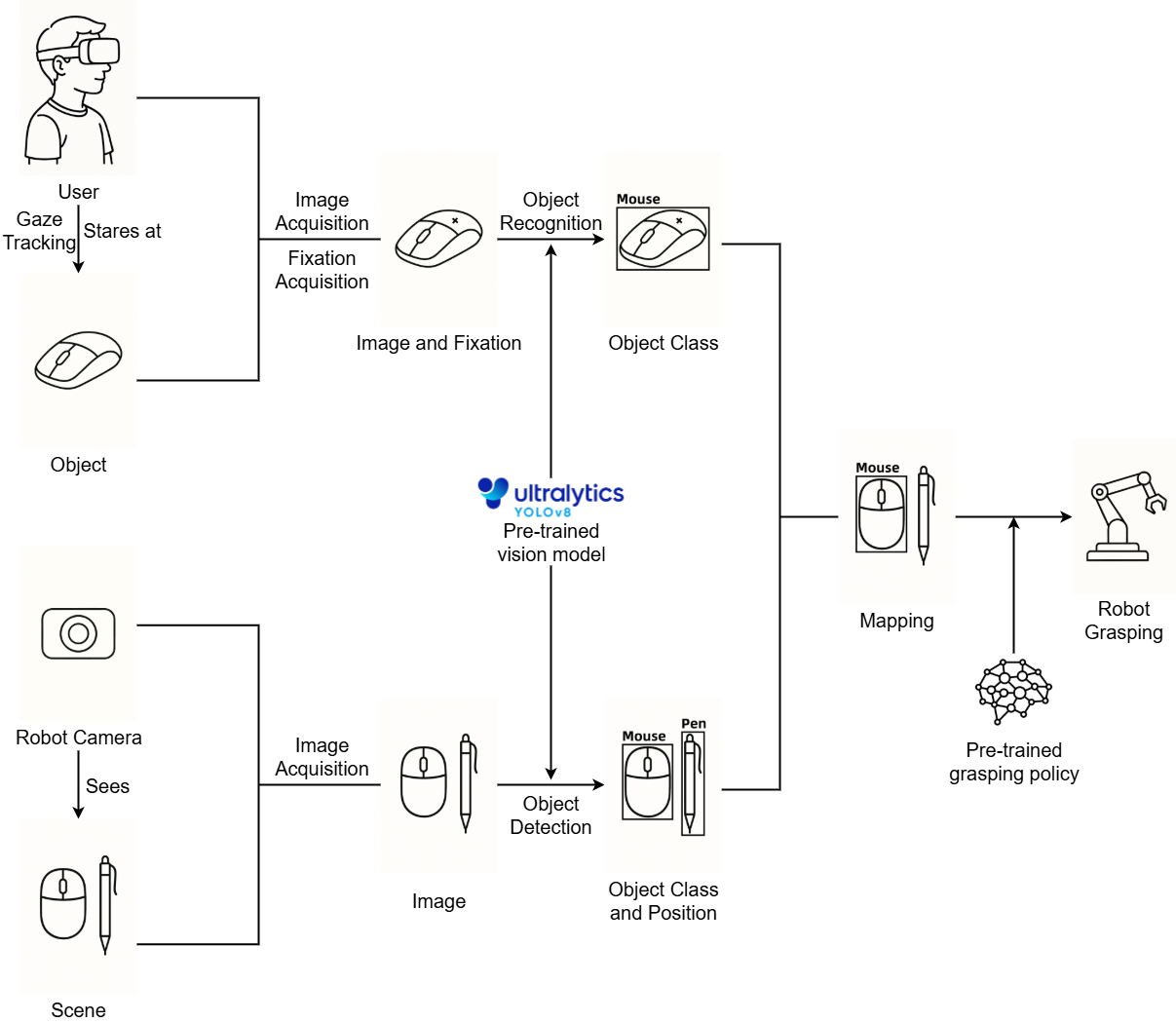} 
    \caption{Overview of the RaycastGrasp system}
    \label{fig:my_label}
\end{figure}

As shown in Figure 1, take grasping a mouse as an example, when a user fixates on an object such as the mouse for a specified duration (gaze time =  2 seconds), the headset automatically captures a screenshot and logs the coordinates obtained through gaze tracking at that moment. YOLOv8 is adopted for object recognition and is used to detect objects in the screenshot captured from the user's perspective. Simultaneously, a camera mounted on the Franka Emika robotic arm captures images from the robot's perspective, which are also processed using the trained YOLOv8 model. The recognized object labels from both perspectives are then mapped to identify a common target. Once the target object is confirmed through label mappings from both the user’s and the robot’s perspectives, the robot executes the corresponding grasping action. This system design bridges the gap between immersive user interaction in MR and practical robotic manipulation, offering a more intuitive, hands-free solution for real-world object handling.

\subsection{Gaze tracking}
To achieve raycast-based eye gaze tracking in a projected passthrough environment, we utilize the embedded eye tracker within the MR headset to capture the user's gaze direction, which is then projected onto a virtual passthrough layer rendered by the headset. This virtual layer is precisely aligned to ensure consistency of gaze data between the virtual and collision spaces. After this alignment, we construct a corresponding collision plane in the scene, positioned immediately after the virtual passthrough layer, which serves as the actual target surface for gaze raycasting, as shown in Figure 2.
\begin{figure}[htbp]
    \centering
    \includegraphics[width=0.8\linewidth]{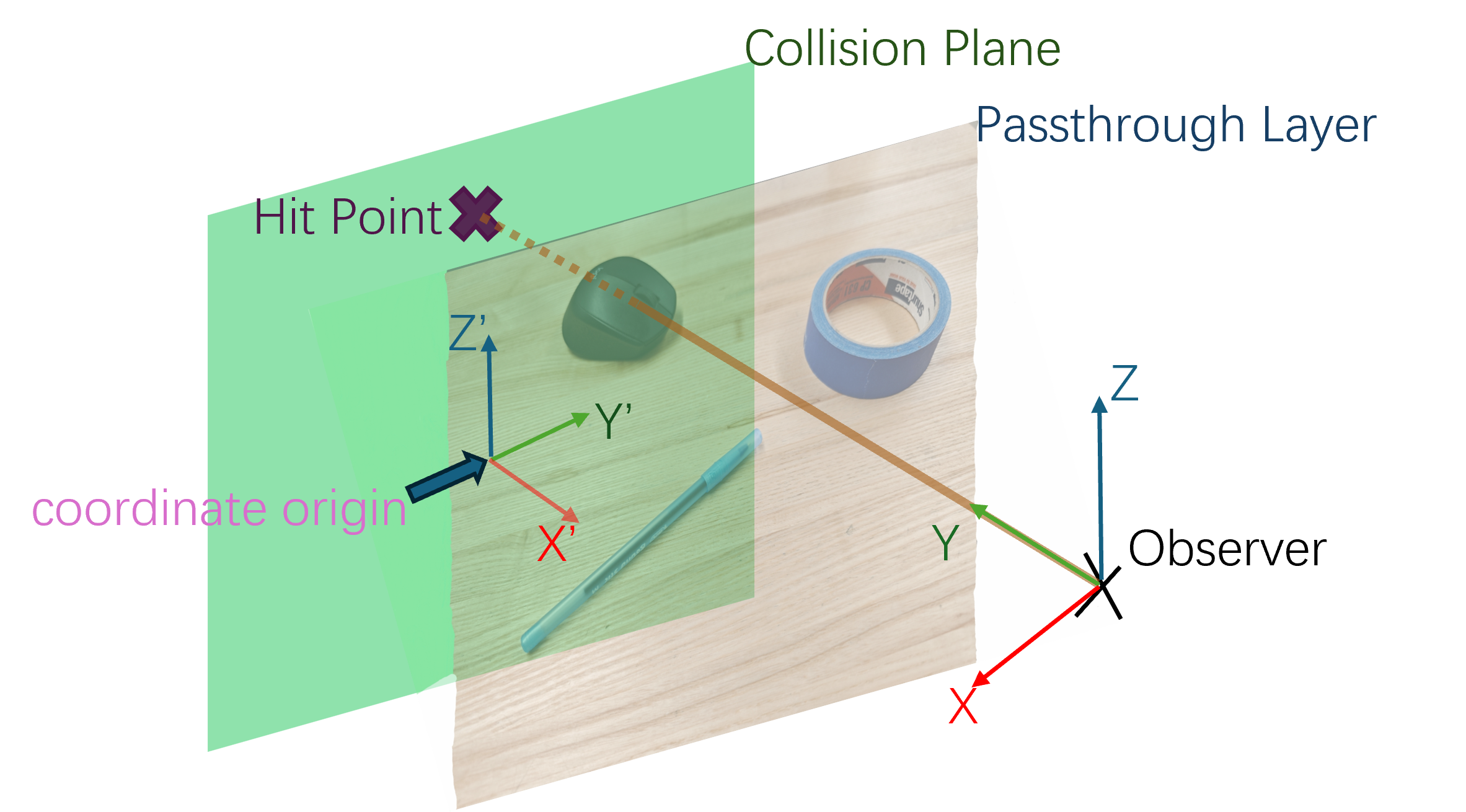} 
    \caption{Alignment of virtual passthrough layer and physical layer}
    \label{fig:my_label}
\end{figure}

To establish an association between gaze targets and semantically recognized real-world objects, we design a workflow that maps eye gaze from Unity’s 3D coordinate space to 2D image space and associates it with YOLO-detected object labels.

When the user focuses on a real-world object through the passthrough layer, the system records the gaze direction as a 3-D normal vector $\textbf{n} \in \mathbb{R}^3$ in Unity’s world space. Using a raycast in the direction \textbf{n}, the 3D gaze point is projected onto a virtual canvas with the hit point $\textbf{x} \in \mathbb{R}^3$, corresponding to a screen pixel coordinate $(u, v), \quad u \in [0, W], \quad v \in [0, H]$ where $H$ and $W$ are the MR screen resolution.

After the screenshot is exported to the host computer, its pixel resolution on disk or during processing may differ from the original screen resolution of the MR device. To ensure coordinate consistency, we rescale the gaze point from the Unity screen space $(u, v)$, defined under the resolution of the MR device $H \times W$, to the coordinate system $\textbf{p} = (u', v')$ of the image on the PC with resolution $H' \times W'$. This transformation is computed as:

\[
u' = \frac{u}{W} \cdot W', \quad v' = \frac{v}{H} \cdot H'
\]

\subsection{Object recognition}

After acquiring the transformed pixel coordinate of the gaze fixation, we will determine the object that the user fixates on. To enable object detection and recognition in the system, we employed the YOLOv8 object detection framework\cite{yolov8}. As a proof of concept, we constructed a simple custom dataset with object labels \textbf{L} by capturing images of common tabletop objects (e.g. mouse, pen, bottle, cup, tape) using a mobile camera.

The data set included two distinct viewpoints to improve detection robustness between perspectives: top-down views simulating the robotic arm’s perspective, and side views approximating the user’s gaze from an MR headset. Each image was manually annotated using the labelImg tool [ref], assigning bounding boxes and class labels to target objects. The images in the dataset contain different objects viewed from different perspectives, allowing the model to generalize between multiple viewing angles. Once trained, the model was integrated into the system to provide localization of objects within the system, supporting gaze-based selection and robotic manipulation.

Each object detected by the trained object recognition model is represented by a rectangular bounding box defined by its top left and bottom right corners in pixel coordinates $[\textbf{p}_1, \textbf{p}_2]$.

To determine whether the user’s gaze intersects any detected object, we check if the transformed gaze pixel coordinate $(u', v')$ lies within any bounding box:
\[
u_1 \leq u' \leq u_2, \quad v_1 \leq v' \leq v_2
\]

If the condition is satisfied, the corresponding label $L_i \in \textbf{L}$ is assigned as the semantic identity of the gaze target:

\[
L_{\text{gaze}} = L_i
\]

This procedure enables semantic grounding of gaze in a real-world MR environment by bridging gaze tracking, coordinate transformation, and object detection.

\subsection{Semantic mapping for robotic manipulation}
The acquired semantic labels will be mapped to the robotic hand-eye camera space for robotic manipulation. The mapping module is designed to align the semantic intention inferred from the user's gaze with the robot's spatial perception of the environment. When a user wearing a VR headset fixates on an object for a duration greater than or equal to two seconds (Time\textsubscript{eye}~$\geq$ 2s), the system automatically captures a screenshot. Simultaneously, a real gaze plane layer (physical layer) positioned behind the device’s passthrough layer records the 3D gaze position of the fixation point. Once both the visual frame and gaze coordinates are obtained, the image is passed to a YOLOv8-based object recognition model, which performs label classification with confidence scores—for example, identifying the object as a “mouse.”

\begin{figure}[htbp]
    \centering
    \includegraphics[width=1\linewidth]{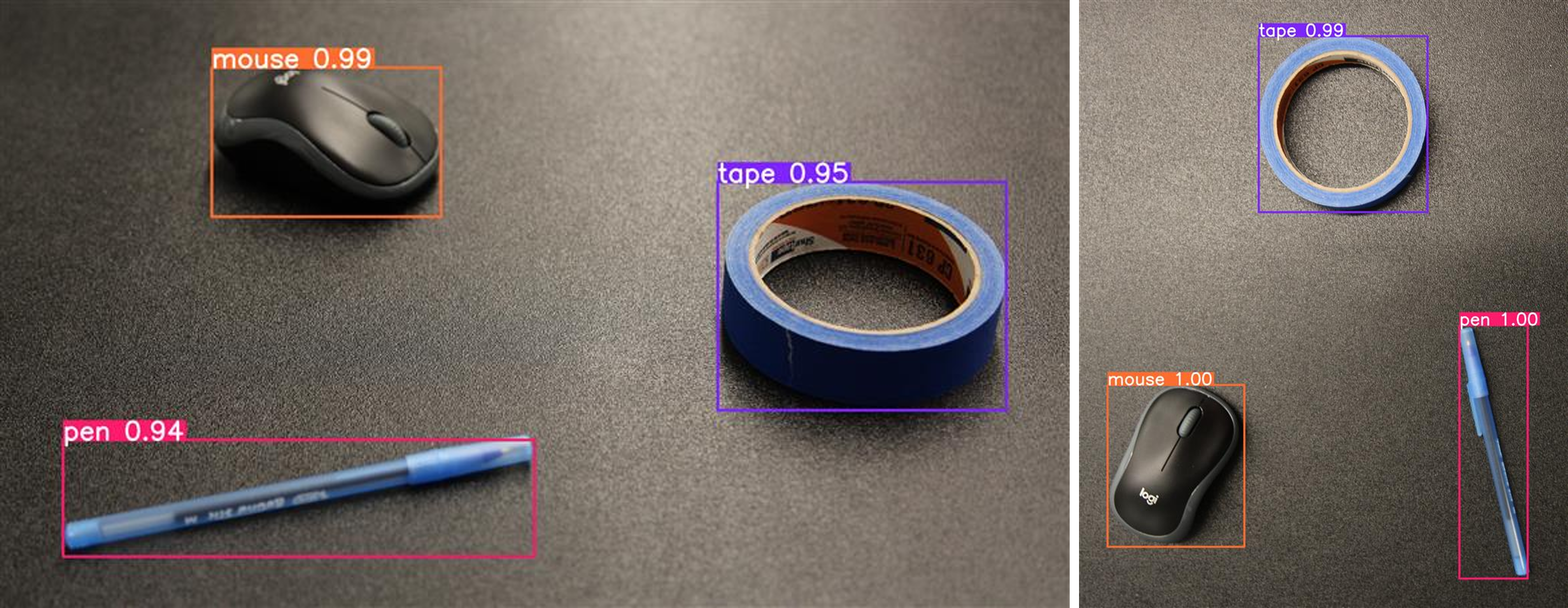} 
    \caption{Class label mapping between user gaze detection and robot camera recognition}
    \label{fig:my_label}
\end{figure}

In parallel, the robot's camera observes the entire workspace and performs object detection, yielding a set of object instances, each associated with a class label (e.g., "mouse", "pen") and positional data. The mapping process compares the user's selected class label with the labels detected in the robot's scene. When a match is found, that is, the label "mouse" exists in the robot’s detected objects, the system determines that this is the object to grasp.

\section{Experiment and Result - demonstration; quantitative results}

\subsection{Experiment Description}

The objective of this experiment is to comprehensively evaluate the performance and user experience of our proposed robotic assistive manipulation system, which is based on a MR headset and gaze tracking technology. As shown in Fig.4. , the experimental setup consists of a Franka Emika collaborative robotic arm, an HTC Focus Vision (MR) headset, a computer configured with both Windows and Linux operating systems, and a rectangular table measuring 60 cm by 90 cm. On the table, common objects such as a water bottle, pen, tape, and mouse are randomly arranged.

\subsection{Demonstration Setup}

As shown in Figures 4 and 5, users are seated in front of the experiment table while wearing the MR headset. Both the headset and the robotic arm are connected to the system host computer to facilitate real-time data transmission and processing.

\begin{figure}[htbp]
    \centering
    \includegraphics[width=1\linewidth]{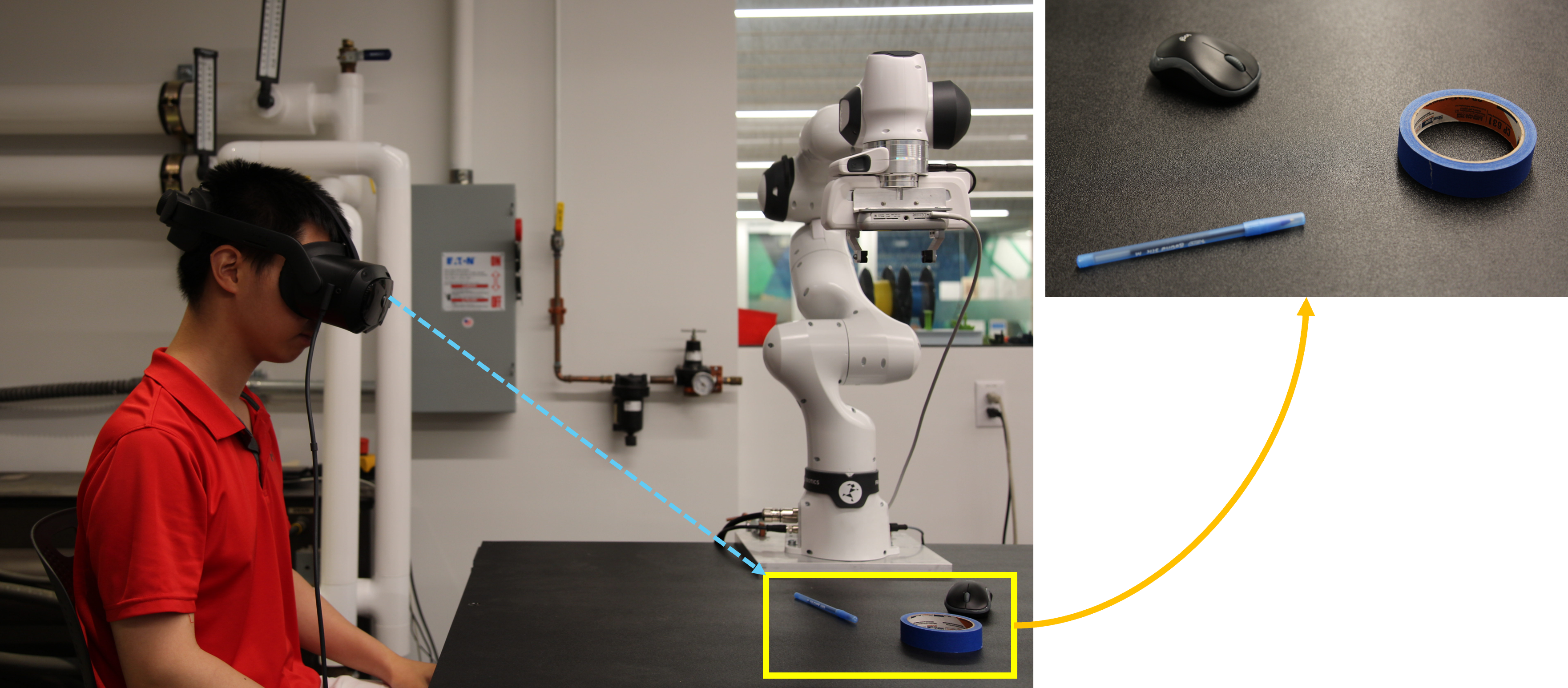} 
    \caption{User observing real objects through VR headset in a mixed reality environment}
    \label{fig:my_label}
\end{figure}

\begin{figure}[htbp]
    \centering
    \includegraphics[width=0.8\linewidth]{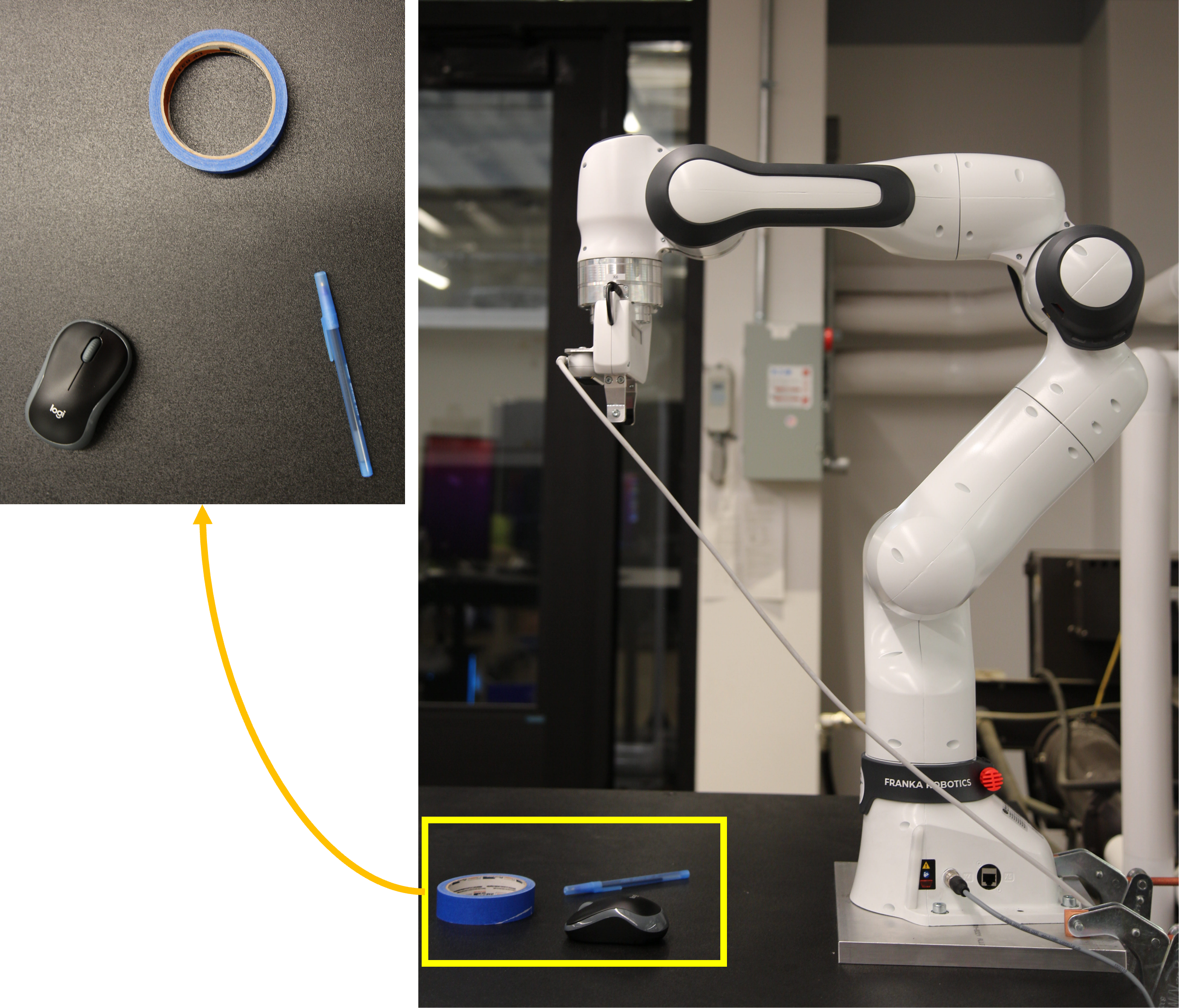} 
    \caption{Object view from the robot-mounted camera perspective}
    \label{fig:my_label}
\end{figure}

\subsection{Experiment Procedure}

During each trial, users are instructed to fixate on any object of their choice on the table. When a participant maintains their gaze on a selected object for more than two seconds and presses a designated button, the system automatically captures a screenshot and records the corresponding gaze coordinates, which are transmitted to the PC. The YOLOv8 object detection model then analyzes the captured image to identify the target object. Based on the recognition result, the Franka Emika robotic arm is instructed to perform a grasping operation on the identified object. Each participant independently completes multiple object selection and grasping tasks, with the entire process monitored and logged by an experimenter.

\subsection{Results}

Experimental results indicate that the gaze tracking system achieves a spatial precision of 0.05 meters. When users maintain their gaze within a region of 0.05 meters for at least two seconds, the system reliably outputs the corresponding coordinates. During image recognition, the YOLOv8 model consistently delivers an object detection confidence $>$ 88\%. The Franka Emika robotic arm demonstrates robust and stable grasping performance, effectively completing the pick-and-place tasks for all tested objects.

\section{Discussion and Conclusion}
The experimental results validate the feasibility and effectiveness of using gaze-based interaction for robotic manipulation in a MR environment. The system achieves over 88\% object detection confidence and demonstrates stable gaze fixation tracking, allowing users to intuitively select and grasp physical objects using only their eye gaze. These findings support our core hypothesis that natural gaze behavior, when combined with real-time object recognition and semantic label mapping, can serve as a reliable control modality for assistive robotic systems.

During the experimental trials, several positive observations were made. First, the system enables a completely hands-free interaction experience, which is particularly critical for users with severe upper-limb impairments. Second, the first-person viewpoint provided by the wearable MR headset enhances user immersion and spatial awareness. Unlike traditional interfaces limited by fixed camera angles, users can interact directly with real-world objects from their natural visual perspective, improving both intuitiveness and usability. Furthermore, the proposed label-based semantic mapping approach simplifies the matching process between user intent and robotic perception, reducing system complexity while maintaining high recognition accuracy.

However, the current mapping strategy exhibits certain limitations. In cluttered or complex environments, multiple instances of the same object class may appear simultaneously, or objects may be positioned too closely, resulting in overlapping bounding boxes. When a user’s gaze point falls within such overlapping regions, or when multiple objects in the robot’s field of view share the same label as the user-intended target, the system may fail to correctly infer the user's true intent. This ambiguity reduces the precision of gaze-to-object mapping and may lead to incorrect grasping decisions. Future research could explore more fine-grained spatial representations, such as instance-level segmentation or temporal gaze pattern analysis, to improve intent disambiguation and enhance system robustness under complex scenarios.

In addition, the system currently uses a fixed 2-second dwell time as the activation threshold. While this static approach helps reduce false positives, it may introduce delays in time-sensitive tasks. To improve responsiveness and interaction flexibility, future work may consider implementing adaptive threshold mechanisms based on gaze stability and contextual awareness.

In conclusion, the RaycastGrasp system advances hands-free, gaze-driven robotic manipulation by leveraging the first-person perspective offered by MR devices. Compared to traditional approaches that rely on interpreting images from robot-mounted cameras for secondary gaze analysis, this system enables direct target selection from the user's natural viewpoint, aligning more closely with intuitive human interaction patterns and enhancing both user experience and operational efficiency. By integrating MR-based interaction with robotic perception through semantic label mapping, the system establishes a lightweight and extensible framework for assistive robotics. Future work will focus on scaling the system to more complex environments, incorporating richer spatial understanding, and enabling robust multi-object disambiguation to improve usability and autonomy.

\vspace{12pt}
\color{black}

\bibliographystyle{IEEEtran}
\bibliography{references}

\end{document}